%% file: main.tex
\documentclass[sigconf]{acmart}
\usepackage{xcolor}
\usepackage{graphicx}
\usepackage{multirow}

\usepackage{tikz}
\newcommand*\circled[1]{\tikz[baseline=(char.base)]{
            \node[shape=circle,draw,inner sep=0.8pt] (char) {#1};}}

\usepackage[ruled, linesnumbered]{algorithm2e}

\copyrightyear{2024}
\acmYear{2024}
\setcopyright{acmlicensed}\acmConference[ISLPED '24]{Proceedings of the ACM/IEEE International Symposium on Low Power Electronics and Design}{August 5--7, 2024}{Newport Beach, CA, USA}
\acmBooktitle{Proceedings of the ACM/IEEE International Symposium on Low Power Electronics and Design (ISLPED '24), August 5--7, 2024, Newport Beach, CA, USA}
\acmDOI{10.1145/3665314.3670831}
\acmISBN{979-8-4007-0688-2/24/08}

\begin{document}


\title{DISHA: Low-Energy Sparse Transformer at Edge for Outdoor Navigation for the Visually Impaired Individuals }
\titlenote{DISHA means "direction" in Hindi and reflects the goal of this work to provide navigational assistance.}

\author{Praveen Nagil, Sumit K. mandal}
\affiliation{%
  \institution{Indian Institute of Science (IISc), Bangalore, India}
  \city{}
  \country{}}


\begin{abstract}

Assistive technology for visually impaired individuals is extremely useful to make them independent of another human being in performing day-to-day chores and instill confidence in them.
One of the important aspects of assistive technology is outdoor navigation for visually impaired people.
While there exist several techniques for outdoor navigation in the literature, they are mainly limited to obstacle detection.
However, navigating a visually impaired person through the sidewalk (while the person is walking outside) is important too.
Moreover, the assistive technology should ensure low-energy operation to extend the battery life of the device.
Therefore, in this work, we propose an end-to-end technology deployed on an edge device to assist visually impaired people.
Specifically, we propose a novel pruning technique for transformer algorithm which detects sidewalk.
The pruning technique ensures low latency of execution and low energy consumption when the pruned transformer algorithm is deployed on the edge device.
Extensive experimental evaluation shows that our proposed technology provides up to 32.49\% improvement in accuracy and 1.4 hours of extension in battery life with respect to a baseline technique.


\end{abstract}

\ccsdesc[500]{Computing methodologies~Video segmentation}
\ccsdesc[500]{Computing methodologies~Scene understanding}

\keywords{Assistive technology, visually impaired, sidewalk detection.}

\maketitle

\input{files/introduction.tex}
\input{files/related_work.tex}

\input{files/background.tex}

\input{files/methodology.tex}
\input{files/experiments.tex}

\input{files/conclusion.tex}

\bibliographystyle{ACM-Reference-Format}
\bibliography{main.bbl}

\end{document}

%% file: files/introduction.tex
\vspace{-2mm}
\section{Introduction}
\vspace{-1mm}

According to the World Health Organization, 40 -- 45 million people are completely visually impaired~\cite{WHO}.
Due to physical limitation, often, visually impaired people lead a lower quality of life which results in deterioration of their mental health~\cite{NIH}.
Many a times people with visual impairment depend on another human being to perform day-to-day tasks.
Constant dependence on another human being may lower the self confidence of the visually impaired people.
One of the important aspect which visually impaired people typically avoid is roaming outdoor on a road shared between vehicles and pedestrians.
Visual impairment may lead to wrong decisions on those roads which may further lead to fatal accidents.

There exist several assistive technology to help visually impaired people to improve their lifestyle.
Most of those assistive technologies are related to indoor and outdoor navigation~\cite{bahadir2012wearable, soveny2014blind,kanwal2015navigation,kang2015novel,kang2017enhanced}.
Kassim et al. proposed a smart cane for visually impaired people.
The cane consists of a device that detects an obstacle on the path and provides feedback through vibration to the person holding the cane if an obstacle is detected~\cite{kassim2016conceptual}.
A wearable sensor-based system is proposed to detect multiple obstacles at the environment while a visually impaired person is walking outdoor~\cite{bahadir2012wearable}.
While obstacle detection is important for outdoor navigation for the visually impaired people, it is not a sufficient aid.
When they walk outdoor, they need to follow the sidewalk (the part of the road dedicated for the pedestrians) so that there is no conflict with the vehicular traffic.
Therefore, we need to build a system which navigates the visually impaired people through the sidewalk as well as detect obstacle and warn them.

State-of-the-art assistive technologies for visually impaired people typically consist of electronic devices with an on-board rechargeable battery.
The electronic devices typically comprise of a set of sensors to collect data and a microcontroller to process the data.
After processing the data, the microcontroller provides the necessary feedback to the user.
However, if the microcontroller performs heavy computations then the on-board battery will be drained soon and it needs to be recharged.
Frequent recharging of the battery is inconvenient especially if the assistive technology is used outdoor and in remote location.
Therefore, apart from correct processing of sensor data, the technology needs to consume low energy to ensure a longer battery life and sustained usage.

To address the aforementioned drawbacks, we propose an end-to-end system with a low-energy and low-storage sparse transformer algorithm for outdoor navigation for visually impaired individuals.
The proposed sparse transformer algorithm takes an outdoor scene as an input image and detects sidewalk and various obstacles present on the image.
We propose a novel platform-aware pruning technique to make the transformer algorithm sparse and energy-efficient.
Specifically, we introduce \textit{k-score} which is a function of number of parameters and latency of execution of different computing block of the transformer algorithm.
According to the \textit{k-score}, we prune a certain fraction of parameters of each computing block of the transformer algorithm to make it energy-efficient.
At the same time, we ensure that the detection accuracy of the transformer algorithm is not degraded by employing activation-based structured pruning. 
Finally, the sparse transformer is deployed on an edge device and it provides voice command to the user to navigate through sidewalk.
Extensive evaluation on synthetic and real outdoor scenarios show that our proposed system extends battery life by up to 1.4 hours while improving the accuracy with respect to a randomly pruned transformer algorithm by up to 32.49\%.
The major contribution of the work are as follows
\vspace{-1mm}
\begin{itemize}
    \item A novel pruning approach for transformer algorithm to improve latency of execution without sacrificing accuracy,
    \item An end-to-end navigation system for visually impaired individuals to help them walk outdoor,
    \item Extensive experimental evaluation showing significant improvement in battery life compared to baseline pruning.
\end{itemize}


%% file: files/related_work.tex
\section{Related Work}

There exist several assistive devices tailored for individuals with visual impairments.
A wearable assistive device is proposed in~\cite{bai2017smart} which generates various sounds to alert users about obstacles ahead, utilizing distinct frequencies and decibel levels to denote obstacle locations.
In~\cite{chuang2018deep}, an autonomous, trail-following robotic guide dog is introduced as a robot-based assistive system.
Furthermore, Wei et al. proposed a guide-dog robot system that utilizes template matching for crossing light and crosswalk detection, aiming to assist visually impaired individuals in achieving independent navigation~\cite{wei2014guide}.
However, these system face difficulty in effectively navigating dynamic environments.

Vision-based assistive systems, developed in~\cite{soveny2014blind,kanwal2015navigation,kang2015novel,kang2017enhanced}, utilize multiple cameras to capture real-world images and execute several computer vision algorithms.
The computer vision algorithms include object detection, segmentation, and image filtering to detect obstacles and provide navigation guidance in different forms. 
Semantic segmentation provides a comprehensive solution to numerous navigational perception problems leading to its swift adoption in visual assistance applications.
Authors in~\cite{bai2019wearable} introduced a device incorporating (RGB-D) cameras, gyroscopes, and smartphones, utilizing segmentation to detect ground and select directions for obstacle navigation indoors and outdoors.
Additionally, a lightweight Convolutional Neural Network (CNN)-based object recognition system on the smartphone is implemented.
Heieh et al. deployed a light CNN-based semantic segmentation network~\cite{hsieh2020outdoor}.
The technology proposed in Fast-SCNN~\cite{poudel2019fast} identifies sidewalk and crosswalk from images captured using an RGB-D depth camera deployed on Nvidia Jetson.
These approaches showcase the utilization of CNN
based segmentation across various applications within the field. 
However, most of the above mentioned techniques rely on third party services (e.g., GPS, internet, cellphone apps) to have the end-to-end system executing at real time.
Availability of third-party services are not guaranteed all the time.
Moreover, a real time assistive systems for visually impaired individuals need to be low-power to sustain a longer battery life and none of the above technique ensures that.

The introduction of transformer algorithms has sparked increased interest among researchers in comparison to traditional CNN-based algorithms for semantic segmentation and object detection.
The introduction of transformers in semantic segmentation, demonstrated by SETR~\cite{zheng2021rethinking}, has shown promising results by utilizing vision transformer (ViT)~\cite{dosovitskiy2020image} as a backbone for feature extraction.
Despite achieving competitive performance on image segmentation benchmarks, transformer-based methods like SETR are limited by their large model size and low efficiency, making real-time deployment challenging and draining the onboard battery of electronic devices quickly.
An alternative approach presented in SegFormer~\cite{xie2021segformer} addresses the concerns of SETR by adopting a hierarchical transformer encoder, which generates both high and low-resolution features and combines them to a lightweight MLP decoder, to produce the final segmentation mask.
By eliminating computationally demanding modules, SegFormer significantly reduces the model size and improves efficiency compared to SETR.
Tan et al.~\cite{tan2021flying} presents a drone-based assistive system prototype utilizing drones to capture the video feed, which is subsequently transmitted to a laptop.
SegFormer gets executed on the laptop to generate segmentation maps and compute direction guidance.
The direction signals are then passed to drone to adjust its location and orientation which are sensed by user through a string connected to drone. 

However, all the existing technologies face significant challenge in terms of energy-efficiency due to their high computation requirements and can drain the battery fast while deployed on edge devices.
To address these challenges we propose a novel pruning technique to produce low-energy sparse transformer algorithm and deployed the transformer on an edge device to construct an end-to-end system for navigating visually impaired individual.
The challenge behind making a transformer algorithm sparse is to maintain the accuracy.
Our proposed platform-aware pruning technique to make the transformer sparse strikes a balance between accuracy and the energy consumption during execution.
Moreover, our proposed system does not require third-party services making the system robust and easy to use.



%% file: files/background.tex
\vspace{-2mm}
\section{Overview of the Proposed End-to-end Navigation System}
\vspace{-1mm}

\begin{figure}[t]
	\centering
	\vspace{-4mm}
	\includegraphics[width=0.48\textwidth]{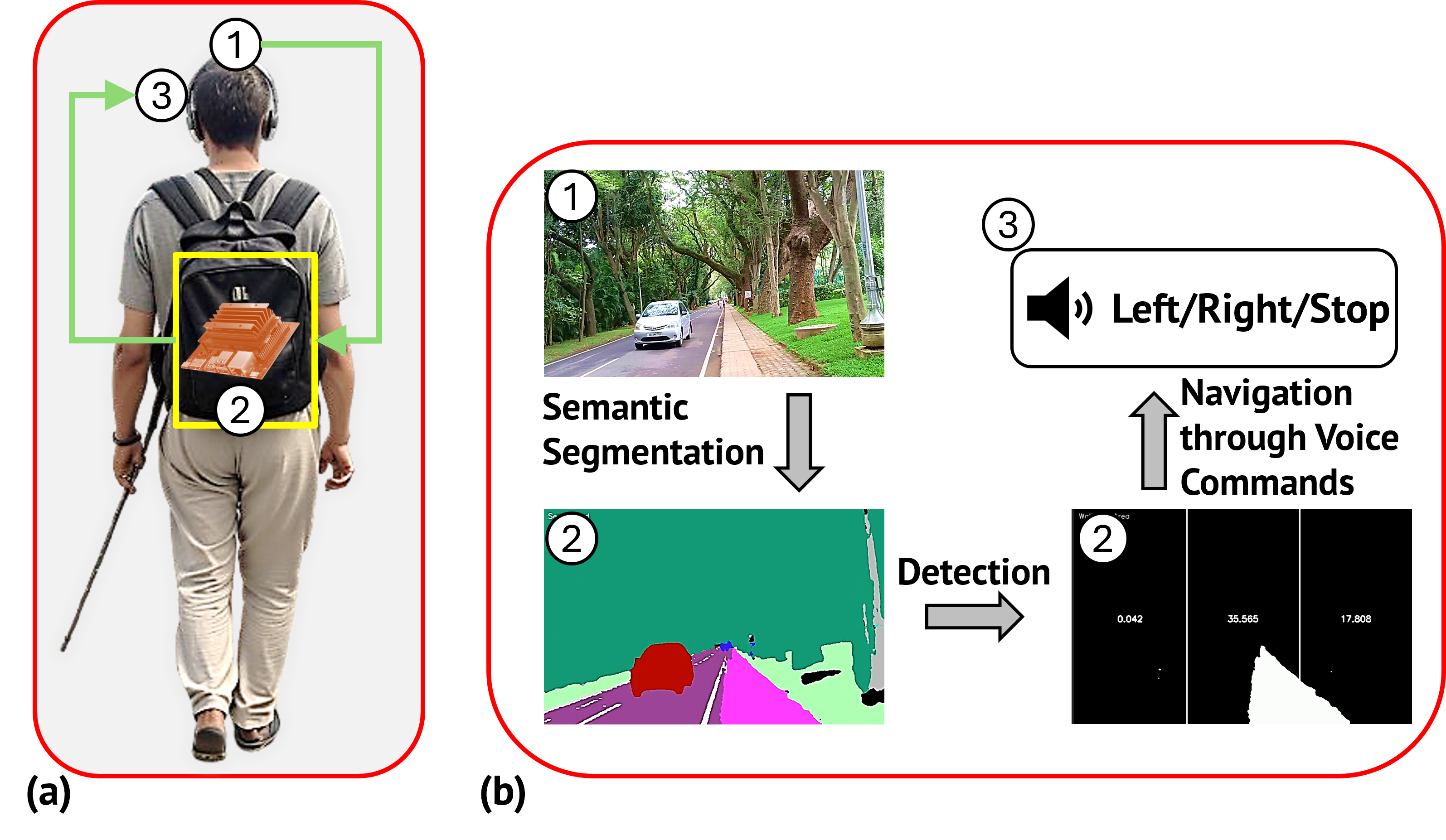}
    \vspace{-4mm}
	\caption{Overview of the Proposed Outdoor Navigation System for Visually Impaired Individuals -- DISHA. (a) shows real time usage and (b) shows the working principle.}
    \label{fig:framework}
    \vspace{-4mm}
\end{figure}

We show the overview of our proposed end-to-end navigation system in Figure~\ref{fig:framework}.
Figure~\ref{fig:framework}(a) shows how the system is used at real time.
The system consists of a camera (marked by \circled{1}), a backpack with an edge device inside (marked by \circled{2}) and a headphone (marked by \circled{3}).
The camera captures the real time video while the user is walking outdoor as shown by (marked by \circled{1}) in Figure~\ref{fig:framework}(b).
Each frame of the video is considered as an image and is processed by the edge device in the backpack.
The processing consists of two steps~--~semantic segmentation and detection.
We propose a low-energy sparse transformer algorithm for the semantic segmentation as described in Section~\ref{sec:method}.
After that, the sidewalk and the obstacles are detected from the segmented output.
The image processing steps are denoted by \circled{2} in Figure~\ref{fig:framework}(b).
Finally, based on the detection, voice commands are sent to the headphone to navigate the user through the sidewalk as shown by (marked by \circled{3}) in the figure.
The details of the implementation are discussed in Section~\ref{sec:end_to_end_impl}.

%% file: files/methodology.tex
\section{Constructing Sparse Transformer Algorithm} \label{sec:method}
\vspace{-1mm}
\subsection{Overview of the Pruning Framework}
\vspace{-1mm}

\begin{figure}[t]
	\centering
	\includegraphics[width=0.45\textwidth]{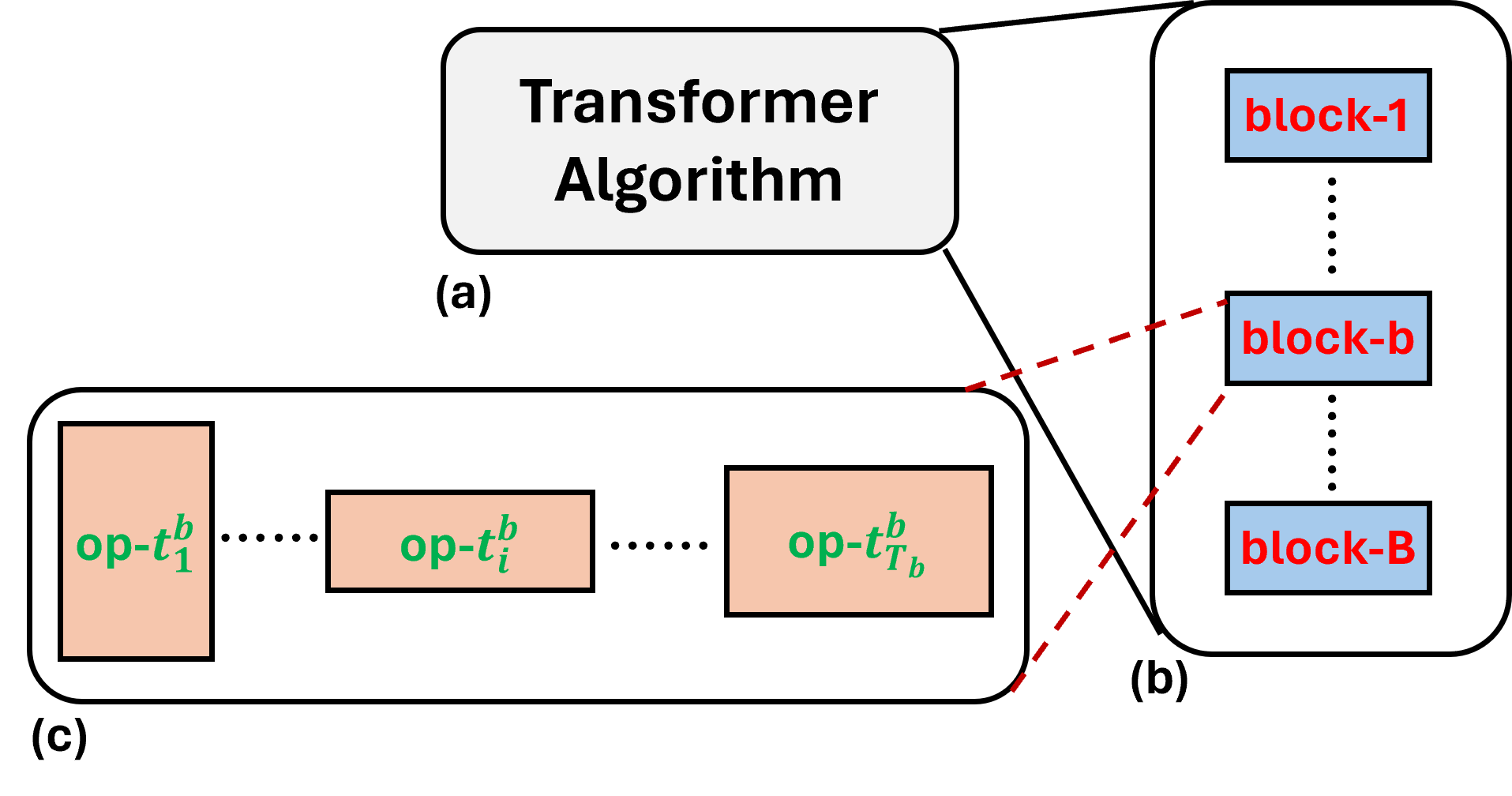} 
	\caption{A transformer algorithm as seen by our proposed pruning framework.}
    \label{fig:transformer}
    \vspace{-5pt}
\end{figure}

Figure~\ref{fig:transformer} shows our representation of a given transformer algorithm used for semantic segmentation.
The proposed pruning technique is general and is applicable to any transformer algorithm.
The input to the transformer algorithm is an outdoor scene (image).
The transformer algorithm segments the image into multiple objects (e.g., car, tree, sidewalk etc.) and classify the objects. 
For any given transformer algorithm, we first divide it in multiple blocks ($b$) as shown in Figure~\ref{fig:transformer}(b).
In this case, we assume that the transformer algorithm consists of total $B$ blocks.
Each block-$b$ consist of a series of operations ($t_i^b$), where $1 \leq i \leq T_b$ and $T_b$ is the total number of operations present in block-$b$ as shown in Figure~\ref{fig:transformer}(c).
Each rectangle in Figure~\ref{fig:transformer}(c) represents an operation and different sizes of the rectangles denote that each operation consists of different input/output size, number of parameters and number of computations. 
We assume that user provides a target pruning ratio ($p$), where $0 < p < 1$.
Our proposed pruning technique consists of two steps.
In the first step, we decide the pruning ratio of each transformer block ($p_b$), where $0 < p_b < 1$.
After that, we decide which parameters to prune for each operations inside each block.
Next, we describe each of these two steps in details.

\subsection{Computing Pruning Ratio ($p_b$) for each Transformer Block}

The goal of our pruning technique is to reduce the size of the transformer algorithm as well as make the transformer algorithm energy-efficient.
Therefore, we first obtain the number of parameters ($w_b$) of each block-$b$, $1 \leq b \leq B$, $B$ is the total number of blocks present in the transformer algorithm.
Since the energy consumption of a transformer algorithm is directly proportional to the inference latency, we perform an offline profiling of the given transformer algorithm on the target edge device to obtain execution latency ($l_b$) of each transformer block.
Next, we define \textit{k-score} ($k_b$) of each transformer block as:
\begin{equation} \label{eq:k_score}
    k_b = w_b \times l_b
\end{equation}
A higher $k_b$ denotes either a higher $w_b$ or a higher $l_b$ or both.
Therefore, the pruning ratio $p_b$ must be proportional to $k_b$ to ensure that the resultant transformer algorithm to be of lower size and energy-efficient.
In other words, the total number of pruned parameters must be distributed among all the transformer blocks based on $k_b$.
A higher number of parameters should get pruned from the block with higher $k_b$.
If $w'$ denotes the total number of pruned parameters from the transformer algorithm and $w_b'$ denotes the number of pruned parameters from each block of the transformer algorithm, then
\begin{equation} \label{eq:prop}
    w_b' \propto k_b \implies w_b' = ck_b
\end{equation}
where $c$ is the proportionality constant.
The total number of pruned parameters $w'$ is expressed as,
\begin{align} \nonumber
    w' & = \sum_{b=1}^B w_b' \\ \nonumber
   \implies w' & = \sum_{b=1}^B ck_b ~~~ (\textrm{substituting} ~w_b' \textrm{~from Equation}~\ref{eq:prop}) \\ \label{eq:c_deriv}
   \implies w' & = c\sum_{b=1}^B k_b = \frac{w'}{\sum_{b=1}^B k_b} 
\end{align}
Substituting Equation~\ref{eq:c_deriv} into Equation~\ref{eq:prop}, we obtain
\begin{equation} \label{eq:final_wb}
    w_b' = w' \frac{k_b}{\sum_{b=1}^B k_b}
\end{equation}
Furthermore, $w'$ and $w_b'$ is expressed as
\begin{equation} \label{eq:pr_exp}
    w' = p \times w, ~~~~~ w_b' = p_b \times w_b
\end{equation}
where $w$ represents the number of parameters present in the original transformer algorithm.
Substituting Equation~\ref{eq:pr_exp} into the Equation~\ref{eq:final_wb}, we obtain
\begin{align} \nonumber
    p_b \times w_b & = p \times w \frac{k_b}{\sum_{b=1}^B k_b} \\ \label{eq:final_pb}
    \implies p_b & = p \frac{w}{w_b} \frac{k_b}{\sum_{b=1}^B k_b} = p \frac{w l_b}{\sum_{b=1}^B k_b}
\end{align}
All the terms in the left hand side of Equation~\ref{eq:final_pb} are known, hence we obtain the desired pruning ratio ($p_b$) for each transformer block.
Next, we discuss the pruning strategy for each transformer block-$b$.

\subsection{Pruning each Transformer Block using $p_b$}

We perform activation-based structured pruning for each transformer block.
As we mentioned earlier, each transformer block consists of a number of operations.
The operations include 2D convolution, linear layer etc.
In our proposed approach, each of these operation inside a transformer block is pruned by a ratio of $p_b$.
We prune the parameters of each operation based on the activation values produced by those parameters.
Lets assume $\mathcal{Z}_i^b$ denotes the set of parameters for the operation $t_i^b$ and $\mathcal{A}_i^b$ denotes the set of sum of absolute of values of the activations produced by each parameter in the set $\mathcal{Z}_i^b$.
We sort the elements in $\mathcal{A}_i^b$ in descending order which produces another set $(\mathcal{A}_i^b)^S$  and then create a set $(\mathcal{Z}_i^b)^S$.
The elements in the set $(\mathcal{Z}_i^b)^S$ correspond to $(\mathcal{A}_i^b)^S$.
Next, we remove $M$ elements from the set $(\mathcal{Z}_i^b)^S$, where $M = p_b \times \big|(\mathcal{Z}_i^b)^S \big|$ and $\big|.\big|$ denotes number of elements in a set.
We obtained the pruned set of parameters after removing those elements.
For convolution operation, we prune desired number of filters keeping the filter size intact.
For linear layer, we prune neurons.
Thus we exploit structured pruning to maximize the performance benefit from pruning.

\begin{figure}[t]
	\centering
	\includegraphics[height=0.3\textwidth]{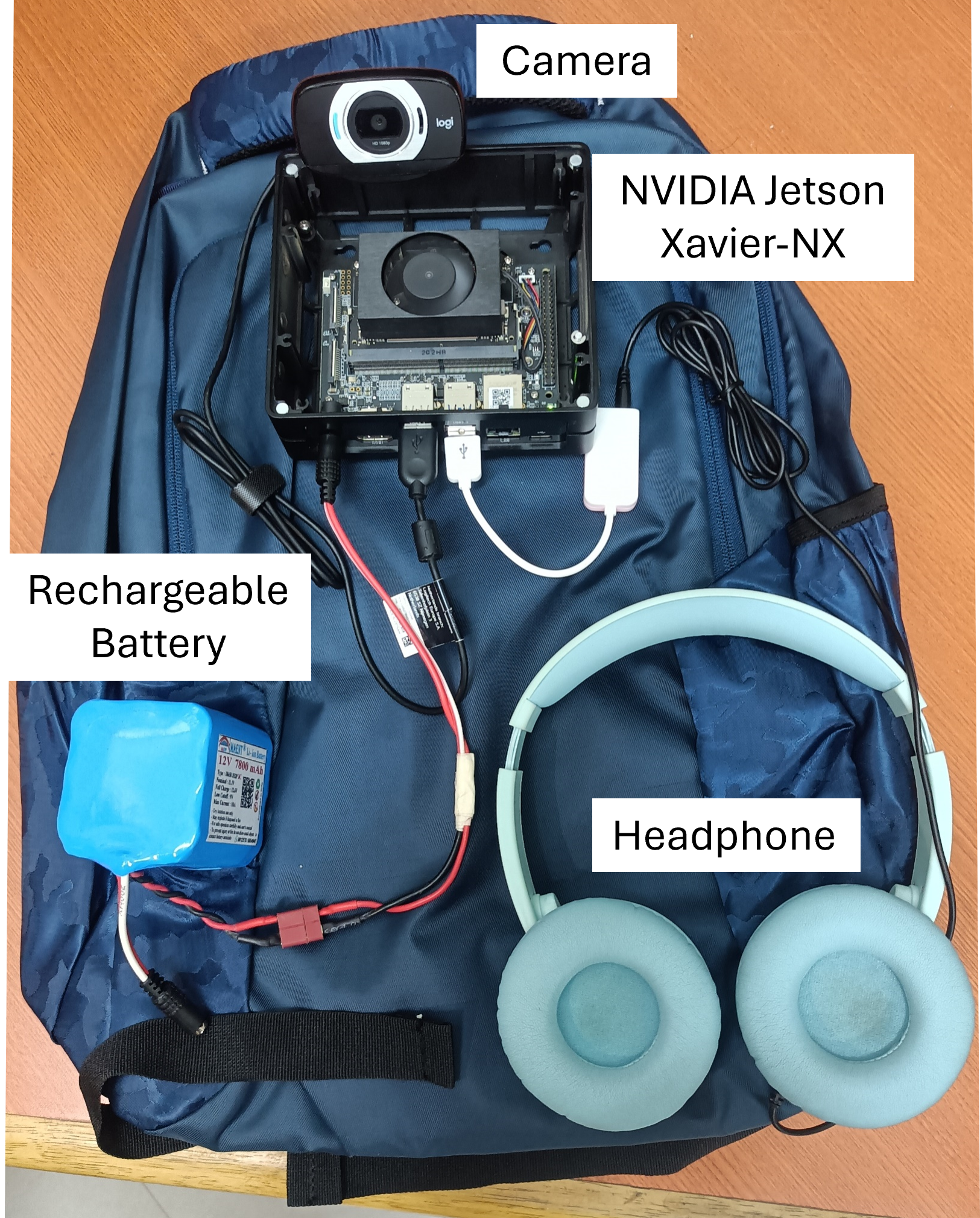} 
	\vspace{-4mm}
	\caption{End-to-end implementation of DISHA.}
    \vspace{-6mm}
	\label{fig:real_system}
\end{figure}

\vspace{-2mm}
\section{End-to-end System Implementation} \label{sec:end_to_end_impl}

We implement the pruned transformer algorithm on NVIDIA Jetson Xavier-NX~\cite{xavier} -- a state-of-the-art edge device.
In this section, we describe different steps of the implementation.
Figure~\ref{fig:real_system} shows the implemented end-to-end system with different components.
The battery and the edge device are put in the backpack.
The camera is mounted on the head of the user, the user wears the backpack and receives the guidance through voice commands via the headphone.

\noindent\textbf{{Execution of the Sparse Transformer Algorithm:}}
The pruned transformer algorithm is integrated into the booting routine of the edge device through a custom startup script executed as a service. Executing the algorithm during the booting ensures that the algorithm initiates automatically upon system boot, enabling it to be \textbf{\textit{operational without manual intervention}}.
A camera is connected as an input to the edge devices and the camera points to the outdoor scene in real time.
During algorithm execution, the system continuously captures frames from the camera and checks for successful frame retrieval.
In case of a failure to capture a frame, appropriate error handling is implemented, accompanied by audio feedback.
Upon capturing a frame, the algorithm applies semantic segmentation to classify pixels into different categories, such as road, sidewalk, buildings, and vehicles.
The resulting segmentation predictions are then processed to identify walkable areas (crosswalks and sidewalks).
These regions are delineated by generating a grayscale mask.
Inspired from an existing approach~\cite{hsieh2020outdoor}, the algorithm divides the image containing the largest walkable area into three horizontal partitions.
Within each partition, a confidence score is computed as the mean of the pixel values within that segment.
We compare these confidence values against a predefined threshold, to establish the reference for direction determination.

\noindent\textbf{{Post-processing the Output from the Algorithm:}}
To enhance the accuracy of directional decisions, we utilize a majority voting mechanism.
Directional choices collected across multiple frames are consolidated using a sliding window approach.
The direction that accumulates the highest count is designated as the ultimate decision.
The sliding window-based technique serves to alleviate transient fluctuations or errors inherent in individual frame processing.
By aggregating directional decisions from multiple frames, the algorithm enhances the reliability of the output direction determination.

\noindent\textbf{{Providing Guidance to the User:}}
Upon determining the direction through majority voting, the algorithm proceeds to generate audio instructions to guide the user effectively.
Leveraging a voice engine (Headphone), the algorithm produces auditory cues aligned with the selected direction.
We deploy five auditory cues -- `Right', `Slight Right', `Left', `Slight Left' and `Stop'.
The integration of audio guidance not only enriches the user experience but also streamlines interaction with the system, fostering seamless communication between the user and the algorithm.

%% file: files/experiments.tex
\vspace{-2mm}
\section{Experimental Evaluation}
\vspace{-1mm}

\subsection{Experimental Setup}

We choose SegFormer~\cite{xie2021segformer} as our base transformer model.
It comes in five versions labeled B0 to B5, with B0 being the baseline one.
We specifically chose B0 because it's smaller in size, hence suitable for deploying in real-time scenarios.
Table~\ref{tab:segformer_arch} shows the structure of SegFormer encoder blocks.
SegFormer comprises of four blocks, each consisting of a combination of linear and convolution operations.
The table describes the types of operations and their respective quantities within each block.

SegFormer performs semantic segmentation of a given outdoor scene and classifies different objects in the scene.
We use two different versions of the transformer algorithm -- one is pre-trained for cityscapes dataset~\cite{cordts2015cityscapes} and another is pre-trained for mapillary vistas dataset~\cite{neuhold2017mapillary}.
While the structure of the transformer algorithm remains same for both of the pre-trained cases, the weight values are different.
The transformer algorithm is pruned for two different input pruning ratios.
We show the comparison in accuracy and the execution time between our proposed pruning technique and random pruning.
Finally, the transformer algorithm is implemented on NVIDIA Jetson Xavier NX and an end-to-end navigation system for visually impaired people is built.
The implementation details of the end-to-end system is described in Section~\ref{sec:end_to_end_impl}.
We note that all experiments are performed on CPU.
Since availability of GPUs are not guaranteed for a given edge device, we do not utilize the available GPUs on the Jetson.

\begin{table}[t]
\centering
\vspace{-4mm}
	\caption{Structure of SegFormer~\cite{xie2021segformer}.}
 \vspace{-2mm}
\label{tab:segformer_arch}
\begin{tabular}{|c|l|l|l}
\cline{1-3}
\multicolumn{1}{|p{1.5cm}|}{SegFormer Encoder} & \multicolumn{1}{p{1.5cm}|}{Types of Operation} & \multicolumn{1}{p{1.5cm}|}{Number of Operations}  & \\ \cline{1-3}
\multirow{2}{*}{block-1}                & Linear             & 4                    &  \\ \cline{2-3}
                                        & Conv               & 10                   &  \\ \cline{1-3}
\multirow{2}{*}{block-2}                & Linear             & 4                    &  \\ \cline{2-3}
                                        & Conv               & 10                   &  \\ \cline{1-3}
\multirow{2}{*}{block-3}                & Linear             & 4                    &  \\ \cline{2-3}
                                        & Conv               & 10                   &  \\ \cline{1-3}
\multirow{2}{*}{block-4}                & Linear             & 2                    &  \\ \cline{2-3}
                                        & Conv               & 10                   &  \\ \cline{1-3}
\end{tabular}
\vspace{-4mm}
\end{table}

\begin{table}[t]
\setlength\tabcolsep{1.5pt}
\centering
\caption{\textit{k-score} and $p_b$ for Different Transformer Blocks.}
 \vspace{-2mm}
\label{tab:k_score}
\begin{tabular}{|l|l|l|l|l|l|}
\hline
\textbf{\begin{tabular}[c]{@{}l@{}}Transformer\\ Block\end{tabular}} & \textbf{\begin{tabular}[c]{@{}l@{}}Latency\\ (ms)\end{tabular}} & \textbf{\begin{tabular}[c]{@{}l@{}}No. of \\ params\end{tabular}} & \textit{\textbf{k-score}} & \textbf{\begin{tabular}[c]{@{}l@{}} $p_b$ for \\ $p=0.35$\end{tabular}} & \textbf{\begin{tabular}[c]{@{}l@{}} $p_b$ for \\ $p=0.40$\end{tabular}} \\ \hline
block-1                                                              & 3.49                 & 158848                                                                & 554062                    &      0.43                   &    0.49                     \\ \hline
block-2                                                              & 3.07                  & 235776                                                                & 723361                    &    0.37                     &    0.43                     \\ \hline
block-3                                                              & 3.57                  & 835200                                                                & 2985005                   &     0.44                    &    0.50                     \\ \hline
block-4                                                              & 2.4                  & 1597952                                                               & 3835085                   &     0.29                    &   0.34                      \\ \hline
\end{tabular}
\end{table}

\subsection{Demonstration of \textit{k-score} and $p_b$ Calculation}

In this section, we show the \textit{k-score} of different blocks in the transformer algorithm.
\textit{k-score} of a transformer block is the product of number of parameters (weights) and the execution time as described in Equation~\ref{eq:k_score}.
Table~\ref{tab:k_score} shows execution time of different transformer blocks of SegFormer for one image inference.
In this case, we took the transformer algorithm pre-trained for cityscapes dataset.
We note that the execution time of different transformer block is not proportional to number of parameters since it also depends on the type of operations (convolution/linear).
\textit{k-score} of different transformer blocks of SegFormer is computed from the execution time and the number of parameters following Equation~\ref{eq:k_score}.
The number of pruned parameter for each transformer block is proportional to its \textit{k-score} according to our technique.
The pruning ratio for transformer block-$b$ ($p_b$) is computed following Equation~\ref{eq:final_pb} and is shown for two input pruning ratio ($p$) in the table.
We observe that $p_b$ is different for each transformer block and depends on its \textit{k-score}.
Next, $p_b$ is used to perform activation-based structured pruning for the transformer blocks.

\begin{figure}[t]
	\centering
	\includegraphics[width=0.48\textwidth]{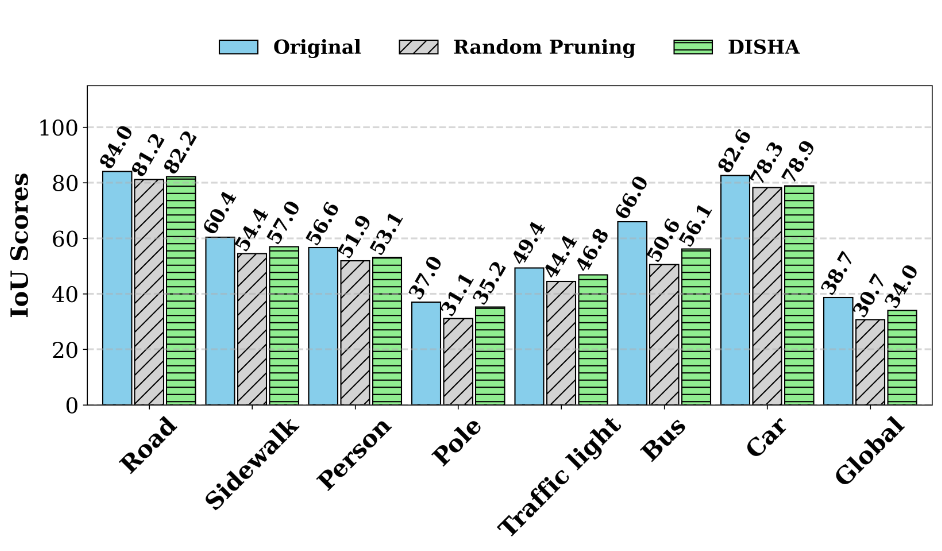} 
	\vspace{-8mm}
	\caption{Comparison of accuracy with \textbf{\textit{mapillary vistas}} dataset with input pruning ratio of \textbf{\textit{0.35}}.}
    \vspace{-4mm}
	\label{fig:map_1_accur}
\end{figure}

\begin{figure}[t]
	\centering
	\includegraphics[width=0.48\textwidth]{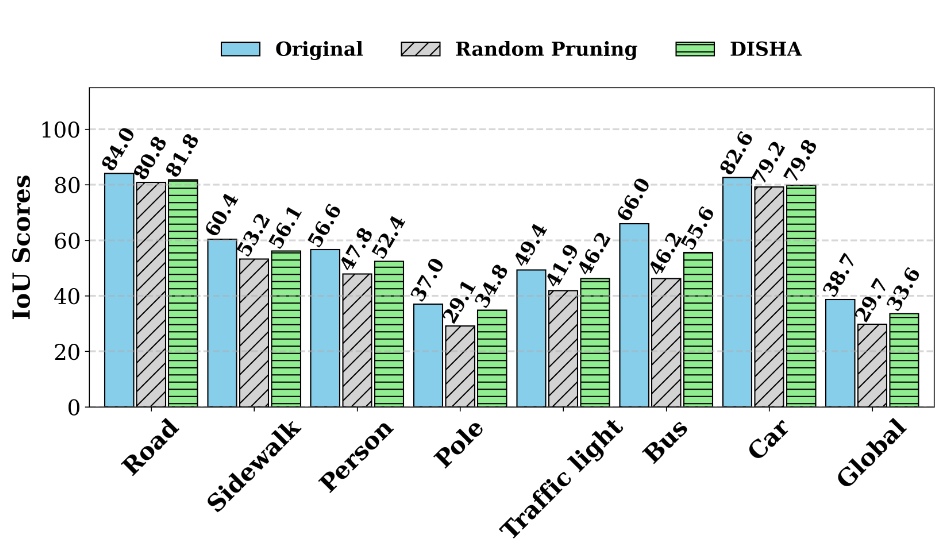} 
	\vspace{-8mm}
	\caption{Comparison of accuracy with \textbf{\textit{mapillary vistas}} dataset with input pruning ratio of \textbf{\textit{0.40}}.}
    \vspace{-4mm}
	\label{fig:map_2_accur}
\end{figure}

\begin{figure}[t]
	\centering
	\includegraphics[width=0.48\textwidth]{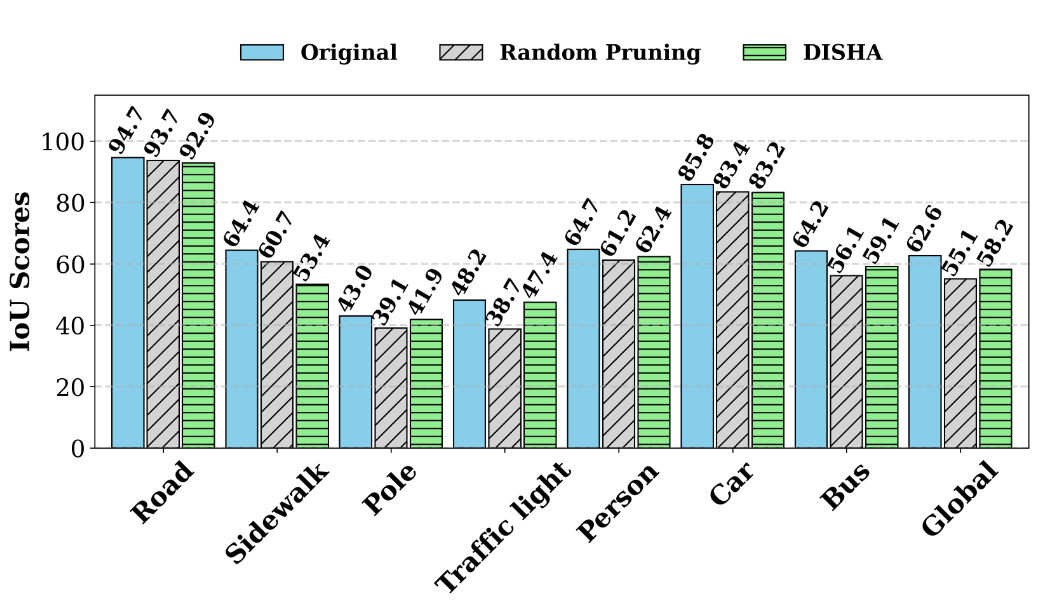} 
	\vspace{-6mm}
	\caption{Comparison of accuracy with \textbf{\textit{cityscapes}} dataset with input pruning ratio of \textbf{\textit{0.35}}.}
    \vspace{-4mm}
	\label{fig:city_1_accur}
\end{figure}

\begin{figure}[t]
	\centering 
	\includegraphics[width=0.48\textwidth]{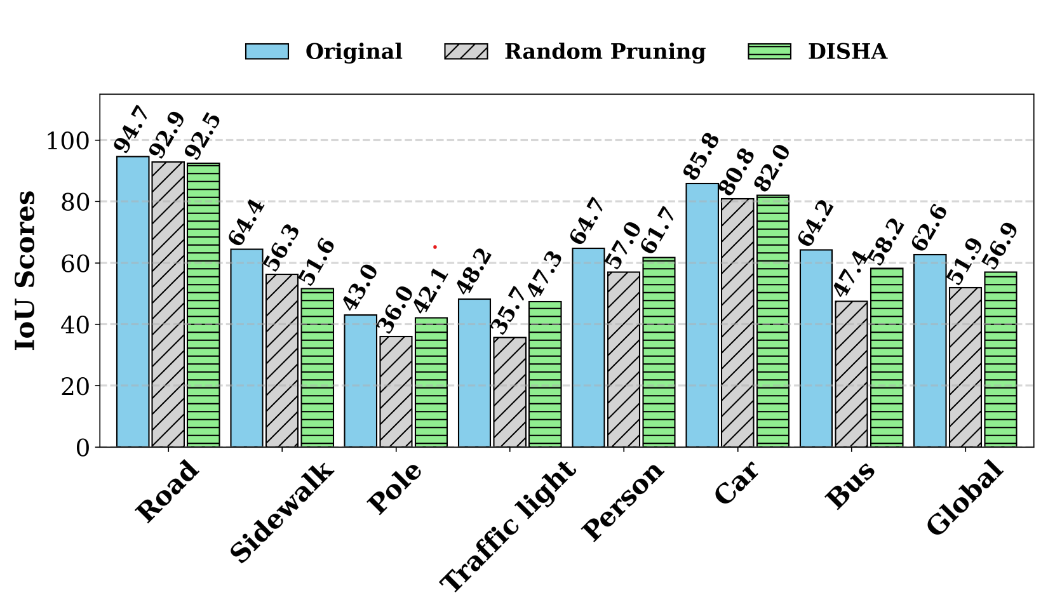}
    \vspace{-6mm}
	\caption{Comparison of accuracy with \textbf{\textit{cityscapes}} dataset with input pruning ratio of \textbf{\textit{0.40}}.}
    \vspace{-4mm}
	\label{fig:city_2_accur}
\end{figure}

\subsection{Comparison in Accuracy}

In this section, we compare the accuracy of the transformer algorithm pruned through our proposed approach (DISHA) with the original algorithm.
We also show the accuracy of a transform algorithm which is randomly pruned with same input pruning ratio ($p$) for each transformer block for a fair comparison.
Figure~\ref{fig:map_1_accur} and Figure~\ref{fig:map_2_accur} shows the comparison for mapillary vistas dataset for pruning ratio of 0.35 and 0.40 respectively.
The y-axis shows the accuracy in terms of intersection of union (IoU) scores which ranges from 0 to 100.
The IoU scores are also mentioned on top of each bar.
The x-axis shows different classes of the dataset.
We only show the results of the classes important for outdoor navigation of visually impaired individuals.
`Global' denotes the average accuracy across all classes.
For both pruning ratio, both of the pruning approach (random pruning and DISHA) achieves lower accuracy than the accuracy of the original transformer algorithm as expected.
However, it is observed that our proposed pruning technique -- DISHA -- improves the accuracy with respect to random pruning technique for all classes.
The improvement in accuracy is computed as:
\begin{equation}
    \textrm{improvement in accuracy} = 100\times \frac{\bigg| \textrm{DISHA} - \textrm{Random} \bigg |}{\textrm{Random}}
\end{equation}
where `DISHA' and `Random' in the above equation denote the accuracy obtained through DISHA and the random pruning approach respectively.
For the pruning ratio of 0.35, the highest improvement of 10.9\% with respect to random pruning is seen for the class of `Bus' and for the pruning ratio of 0.40, the highest improvement of 19.59\% is seen for the class of `Pole'.
Overall, DISHA improves the accuracy by 10.75\% and 13.13\% for the pruning ratio of 0.35 and 0.40 respectively compared to the random pruning approach.

We observe similar result for cityscape dataset too.
The results for cityscape dataset is shown in Figure~\ref{fig:city_1_accur} and Figure~\ref{fig:city_2_accur} for pruning ratio of 0.35 and 0.40 respectively.
For the pruning ratio of 0.35, the highest improvement of 22.49\% with respect to random pruning is seen for the class of `Traffic Light' and for the pruning ratio of 0.40, the highest improvement of 32.49\% is seen for the class of `Traffic light' too.
Overall, DISHA improves the accuracy by 5.63\% and 9.63\% for the pruning ratio of 0.35 and 0.40 respectively compared to the random pruning approach.
The impressive improvement originates from our activation-based pruning.
Since we choose the parameters which produce lowest activations (in terms of absolute values), the effect of pruning is less on the accuracy compared to random pruning.

\begin{table}[t]
\caption{Percentage reduction in latency and energy for different pruning techniques with different pruning ratios.}
\vspace{-4mm}
	\label{tab:latency_energy}
\begin{tabular}{|l|ll||ll|}
\hline
\multirow{2}{*}{} & \multicolumn{2}{c||}{Latency (\%)} & \multicolumn{2}{c|}{Energy (\%)} \\ \cline{2-5} 
                  & \multicolumn{1}{l|}{Random}    & DISHA    & \multicolumn{1}{l|}{Random}    & DISHA   \\ \hline
$p=0.35$          & \multicolumn{1}{l|}{13.63}          &  17.75        & \multicolumn{1}{l|}{15.61}          &  19.16       \\ \hline
$p=0.40$          & \multicolumn{1}{l|}{15.83}         &  19.08       & \multicolumn{1}{l|}{17.19}          & 20.46        \\ \hline
\end{tabular}
\vspace{-4mm}
	\end{table}

\vspace{-2mm}
\subsection{Improvement in Execution Latency and Energy Consumption}

Table~\ref{tab:latency_energy} shows the percentage reduction in latency and energy of execution for random pruning and DISHA with respect to the transformer algorithm without pruning.
We observe that for $p=0.35$, random pruning technique reduces the latency by 13.63\% while DISHA reduces the latency by 17.75\%.
Therefore, DISHA reduces 4.12\% more latency of execution.
Similarly, DISHA reduces 3.25\% more latency of execution for $p=0.40$ with respect to random pruning.
Since number of parameters pruned with DISHA in each transformer block is proportional to their \textit{k-score} (Equation~\ref{eq:prop}) and \textit{k-score} is a monotonically increasing function of latency of each transformer block, we observe more reduction in latency with DISHA compared to the random pruning approach.

Improvement in latency further results in improvement in energy consumption.
We obtain average power consumption for each transformer block and use it to compute the energy of execution.
Table~\ref{tab:latency_energy} reveals that DISHA improves the reduction in energy consumption by 3.55\% and 3.27\% for $p=0.35$ and $p=0.40$ respectively.
The improvement in latency and energy of execution results in approximately \textbf{\textit{1.4 hours of more battery life}} with a 7800 mAh 12V battery (with 9V as cutoff).


%% file: files/conclusion.tex
\vspace{-2mm}
\section{Conclusion}
\vspace{-1mm}
	
Assistive technology for visually impaired individual is important to move them one step towards self-sufficiency.
There exists several assistive technology for visually impaired people which detects obstacle outdoor.
While obstacle detection is necessary for navigating people, it is not sufficient.
The assistive technology should also detect sidewalk from an outdoor scene and navigate the visually impaired person.
Moreover, the technology should be energy-efficient to avoid frequent recharge of the on-board battery.
To this end, we construct an end-to-end system which navigates the visually impaired individuals outdoor.
To make the system energy-efficient we propose a novel pruning approach for a transformer algorithm.
The transformer algorithm segments an outdoor scene into multiple objects and detects sidewalk.
Extensive experimental evaluation shows that our proposed technique extends battery life by 1.4 hours while improving the accuracy of detection by up to 32.49\% with respect to a random pruning approach.

\vspace{-2mm}

\section*{Acknowledgement}
This work is supported in part by Intel Technology India Private Limited and the Walmart Center for Tech Excellence at IISc (CSR Grant WMGT-23-0001).

\vspace{-2mm}

%% file: main.bbl

\begin{thebibliography}{21}


\ifx \showCODEN    \undefined \def \showCODEN     #1{\unskip}     \fi
\ifx \showDOI      \undefined \def \showDOI       #1{#1}\fi
\ifx \showISBNx    \undefined \def \showISBNx     #1{\unskip}     \fi
\ifx \showISBNxiii \undefined \def \showISBNxiii  #1{\unskip}     \fi
\ifx \showISSN     \undefined \def \showISSN      #1{\unskip}     \fi
\ifx \showLCCN     \undefined \def \showLCCN      #1{\unskip}     \fi
\ifx \shownote     \undefined \def \shownote      #1{#1}          \fi
\ifx \showarticletitle \undefined \def \showarticletitle #1{#1}   \fi
\ifx \showURL      \undefined \def \showURL       {\relax}        \fi
\providecommand\bibfield[2]{#2}
\providecommand\bibinfo[2]{#2}
\providecommand\natexlab[1]{#1}
\providecommand\showeprint[2][]{arXiv:#2}

\bibitem[Bahadir et~al\mbox{.}(2012)]%
        {bahadir2012wearable}
\bibfield{author}{\bibinfo{person}{Senem~Kursun Bahadir} {et~al\mbox{.}}} \bibinfo{year}{2012}\natexlab{}.
\newblock \showarticletitle{{Wearable Obstacle Detection System fully Integrated to Textile Structures for Visually Impaired People}}.
\newblock \bibinfo{journal}{\emph{Sensors and Actuators A: Physical}}  \bibinfo{volume}{179} (\bibinfo{year}{2012}), \bibinfo{pages}{297--311}.
\newblock


\bibitem[Bai et~al\mbox{.}(2017)]%
        {bai2017smart}
\bibfield{author}{\bibinfo{person}{Jinqiang Bai} {et~al\mbox{.}}} \bibinfo{year}{2017}\natexlab{}.
\newblock \showarticletitle{Smart guiding glasses for visually impaired people in indoor environment}.
\newblock \bibinfo{journal}{\emph{IEEE Trans. on Consumer Electronics}} \bibinfo{volume}{63}, \bibinfo{number}{3} (\bibinfo{year}{2017}), \bibinfo{pages}{258--266}.
\newblock


\bibitem[Bai et~al\mbox{.}(2019)]%
        {bai2019wearable}
\bibfield{author}{\bibinfo{person}{Jinqiang Bai} {et~al\mbox{.}}} \bibinfo{year}{2019}\natexlab{}.
\newblock \showarticletitle{Wearable travel aid for environment perception and navigation of visually impaired people}.
\newblock \bibinfo{journal}{\emph{Electronics}} \bibinfo{volume}{8}, \bibinfo{number}{6} (\bibinfo{year}{2019}), \bibinfo{pages}{697}.
\newblock


\bibitem[Chuang et~al\mbox{.}(2018)]%
        {chuang2018deep}
\bibfield{author}{\bibinfo{person}{Tzu-Kuan Chuang} {et~al\mbox{.}}} \bibinfo{year}{2018}\natexlab{}.
\newblock \showarticletitle{Deep trail-following robotic guide dog in pedestrian environments for people who are blind and visually impaired-learning from virtual and real worlds}. In \bibinfo{booktitle}{\emph{2018 IEEE ICRA}}. \bibinfo{pages}{5849--5855}.
\newblock


\bibitem[Cordts et~al\mbox{.}(2015)]%
        {cordts2015cityscapes}
\bibfield{author}{\bibinfo{person}{Marius Cordts} {et~al\mbox{.}}} \bibinfo{year}{2015}\natexlab{}.
\newblock \showarticletitle{{The Cityscapes Dataset}}. In \bibinfo{booktitle}{\emph{CVPR Workshop on the Future of Datasets in Vision}}, Vol.~\bibinfo{volume}{2}. \bibinfo{pages}{1}.
\newblock


\bibitem[Dosovitskiy et~al\mbox{.}(2020)]%
        {dosovitskiy2020image}
\bibfield{author}{\bibinfo{person}{Alexey Dosovitskiy} {et~al\mbox{.}}} \bibinfo{year}{2020}\natexlab{}.
\newblock \showarticletitle{An image is worth 16x16 words: Transformers for image recognition at scale}.
\newblock \bibinfo{journal}{\emph{arXiv preprint arXiv:2010.11929}} (\bibinfo{year}{2020}).
\newblock


\bibitem[Hsieh et~al\mbox{.}(2020)]%
        {hsieh2020outdoor}
\bibfield{author}{\bibinfo{person}{I-Hsuan Hsieh} {et~al\mbox{.}}} \bibinfo{year}{2020}\natexlab{}.
\newblock \showarticletitle{Outdoor walking guide for the visually-impaired people based on semantic segmentation and depth map}. In \bibinfo{booktitle}{\emph{2020 international conference on pervasive artificial intelligence (ICPAI)}}. IEEE, \bibinfo{pages}{144--147}.
\newblock


\bibitem[Kang et~al\mbox{.}(2015)]%
        {kang2015novel}
\bibfield{author}{\bibinfo{person}{Mun-Cheon Kang} {et~al\mbox{.}}} \bibinfo{year}{2015}\natexlab{}.
\newblock \showarticletitle{A novel obstacle detection method based on deformable grid for the visually impaired}.
\newblock \bibinfo{journal}{\emph{IEEE Transactions on Consumer Electronics}} \bibinfo{volume}{61}, \bibinfo{number}{3} (\bibinfo{year}{2015}), \bibinfo{pages}{376--383}.
\newblock


\bibitem[Kang et~al\mbox{.}(2017)]%
        {kang2017enhanced}
\bibfield{author}{\bibinfo{person}{Mun-Cheon Kang} {et~al\mbox{.}}} \bibinfo{year}{2017}\natexlab{}.
\newblock \showarticletitle{An enhanced obstacle avoidance method for the visually impaired using deformable grid}.
\newblock \bibinfo{journal}{\emph{IEEE Transactions on Consumer Electronics}} \bibinfo{volume}{63}, \bibinfo{number}{2} (\bibinfo{year}{2017}), \bibinfo{pages}{169--177}.
\newblock


\bibitem[Kanwal et~al\mbox{.}(2015)]%
        {kanwal2015navigation}
\bibfield{author}{\bibinfo{person}{Nadia Kanwal} {et~al\mbox{.}}} \bibinfo{year}{2015}\natexlab{}.
\newblock \showarticletitle{A navigation system for the visually impaired: a fusion of vision and depth sensor}.
\newblock \bibinfo{journal}{\emph{Applied bionics and biomechanics}}  \bibinfo{volume}{2015} (\bibinfo{year}{2015}).
\newblock


\bibitem[Kassim et~al\mbox{.}(2016)]%
        {kassim2016conceptual}
\bibfield{author}{\bibinfo{person}{Anuar~Mohamed Kassim} {et~al\mbox{.}}} \bibinfo{year}{2016}\natexlab{}.
\newblock \showarticletitle{{Conceptual Design and Implementation of Electronic Spectacle based Obstacle Detection for Visually Impaired Persons}}.
\newblock \bibinfo{journal}{\emph{Journal of Advanced Mechanical Design, Systems, and Manufacturing}} \bibinfo{volume}{10}, \bibinfo{number}{7} (\bibinfo{year}{2016}), \bibinfo{pages}{JAMDSM0094--JAMDSM0094}.
\newblock


\bibitem[Neuhold et~al\mbox{.}(2017)]%
        {neuhold2017mapillary}
\bibfield{author}{\bibinfo{person}{Gerhard Neuhold} {et~al\mbox{.}}} \bibinfo{year}{2017}\natexlab{}.
\newblock \showarticletitle{{The Mapillary Vistas Dataset for Semantic understanding of Street Scenes}}. In \bibinfo{booktitle}{\emph{Proceedings of the IEEE international conference on computer vision}}. \bibinfo{pages}{4990--4999}.
\newblock


\bibitem[NIH({[n.\,d.]})]%
        {NIH}
\bibfield{author}{\bibinfo{person}{NIH}.} \bibinfo{year}{[n.\,d.]}\natexlab{}.
\newblock \bibinfo{booktitle}{\emph{{Visual Impairment and Mental Health: Unmet Needs and Treatment Options}}}.
\newblock
\newblock
\shownote{\url{https://www.ncbi.nlm.nih.gov/pmc/articles/PMC7721280/}, accessed 8 March 2024.}.


\bibitem[NVIDIA({[n.\,d.]})]%
        {xavier}
\bibfield{author}{\bibinfo{person}{NVIDIA}.} \bibinfo{year}{[n.\,d.]}\natexlab{}.
\newblock \bibinfo{booktitle}{\emph{{Jetson Xavier NX}}}.
\newblock
\newblock
\shownote{\url{https://www.nvidia.com/en-in/autonomous-machines/embedded-systems/jetson-xavier-nx/}, accessed 26 July 2023.}.


\bibitem[Poudel et~al\mbox{.}(2019)]%
        {poudel2019fast}
\bibfield{author}{\bibinfo{person}{Rudra~PK Poudel}, \bibinfo{person}{Stephan Liwicki}, {and} \bibinfo{person}{Roberto Cipolla}.} \bibinfo{year}{2019}\natexlab{}.
\newblock \showarticletitle{Fast-scnn: Fast semantic segmentation network}.
\newblock \bibinfo{journal}{\emph{arXiv preprint arXiv:1902.04502}} (\bibinfo{year}{2019}).
\newblock


\bibitem[S{\"o}veny et~al\mbox{.}(2014)]%
        {soveny2014blind}
\bibfield{author}{\bibinfo{person}{B{\'a}lint S{\"o}veny}, \bibinfo{person}{G{\'a}bor Kov{\'a}cs}, {and} \bibinfo{person}{Zsolt~T Kardkov{\'a}cs}.} \bibinfo{year}{2014}\natexlab{}.
\newblock \showarticletitle{Blind guide-A virtual eye for guiding indoor and outdoor movement}. In \bibinfo{booktitle}{\emph{2014 5th IEEE Conference on Cognitive Infocommunications (CogInfoCom)}}. IEEE, \bibinfo{pages}{343--347}.
\newblock


\bibitem[Tan et~al\mbox{.}(2021)]%
        {tan2021flying}
\bibfield{author}{\bibinfo{person}{Haobin Tan} {et~al\mbox{.}}} \bibinfo{year}{2021}\natexlab{}.
\newblock \showarticletitle{Flying guide dog: Walkable path discovery for the visually impaired utilizing drones and transformer-based semantic segmentation}. In \bibinfo{booktitle}{\emph{2021 IEEE ROBIO}}. \bibinfo{pages}{1123--1128}.
\newblock


\bibitem[Wei and Lee(2014)]%
        {wei2014guide}
\bibfield{author}{\bibinfo{person}{Yuanlong Wei} {and} \bibinfo{person}{Mincheol Lee}.} \bibinfo{year}{2014}\natexlab{}.
\newblock \showarticletitle{A guide-dog robot system research for the visually impaired}. In \bibinfo{booktitle}{\emph{2014 IEEE ICIT}}. \bibinfo{pages}{800--805}.
\newblock


\bibitem[WHO({[n.\,d.]})]%
        {WHO}
\bibfield{author}{\bibinfo{person}{WHO}.} \bibinfo{year}{[n.\,d.]}\natexlab{}.
\newblock \bibinfo{booktitle}{\emph{{Up to 45 million blind people globally - and growing}}}.
\newblock
\newblock
\shownote{\url{https://tinyurl.com/2uta3fcy}, accessed 8 March 2024.}.


\bibitem[Xie et~al\mbox{.}(2021)]%
        {xie2021segformer}
\bibfield{author}{\bibinfo{person}{Enze Xie} {et~al\mbox{.}}} \bibinfo{year}{2021}\natexlab{}.
\newblock \showarticletitle{{SegFormer: Simple and Efficient Design for Semantic Segmentation with Transformers}}.
\newblock \bibinfo{journal}{\emph{NeuRIPS}}  \bibinfo{volume}{34} (\bibinfo{year}{2021}), \bibinfo{pages}{12077--12090}.
\newblock


\bibitem[Zheng et~al\mbox{.}(2021)]%
        {zheng2021rethinking}
\bibfield{author}{\bibinfo{person}{Sixiao Zheng} {et~al\mbox{.}}} \bibinfo{year}{2021}\natexlab{}.
\newblock \showarticletitle{Rethinking semantic segmentation from a sequence-to-sequence perspective with transformers}. In \bibinfo{booktitle}{\emph{Proc. of CVPR}}. \bibinfo{pages}{6881--6890}.
\newblock


\end{thebibliography}
